\documentclass[]{llncs}
\usepackage{graphicx}
\usepackage{times}
\usepackage{soul}
\usepackage{url}
\usepackage[hidelinks]{hyperref}
\usepackage[utf8]{inputenc}
\usepackage{amsmath}
\usepackage{booktabs}
\usepackage[marginal]{footmisc}
\usepackage{CJK}
\usepackage[ruled,linesnumbered]{algorithm2e}
\usepackage{amsfonts,amssymb}
\usepackage{enumitem}
\usepackage{subfigure}
\usepackage{makecell}
\usepackage{multirow}
\usepackage{bm}
\usepackage{hyperref}
\usepackage{times}
\usepackage{tabularx}
\usepackage{authblk}
\usepackage{xcolor}

\begin{document}
\title{Probing Negative Sampling Strategies to Learn Graph Representations via Unsupervised Contrastive Learning}

% \author{}
% \authorrunning{}
% \institute{}

%
%\titlerunning{Abbreviated paper title}
% If the paper title is too long for the running head, you can set
% an abbreviated paper title here
%
\author{Shiyi~Chen\textsuperscript{1} \and
Ziao~Wang\textsuperscript{2}\and
Xinni~Zhang\textsuperscript{3}\and
Xiaofeng~Zhang\textsuperscript{4} \footnote{Contact Author}\and
Dan~Peng\textsuperscript{5}}

% %
% \authorrunning{F. Author et al.}
% % First names are abbreviated in the running head.
% % If there are more than two authors, 'et al.' is used.
% %
\institute{Harbin Institute of Technology (Shenzhen), China\\
\email{\{19s151083\textsuperscript{1},20s051054\textsuperscript{3},pengdan\textsuperscript{5}\}@stu.hit.edu.cn,\
wangziao1993\textsuperscript{2}@hotmail.com,zhangxiaofeng\textsuperscript{4}@hit.edu.cn}}

% %

\maketitle  
\begin{abstract}
Graph representation learning has long been an important yet challenging task for various real-world applications. However, their downstream tasks are mainly performed in the settings of supervised or semi-supervised learning. %focus on node classification, link prediction and graph classification under semi-supervised learning. 
Inspired by recent advances in unsupervised contrastive learning, % from computer vision and natural language processing, 
%we propose a novel node-wise contrastive learning approach.  
this paper is thus motivated to investigate how the node-wise contrastive learning could be performed. %Particularly, we assume the positive instance propose to transit the possible positive instances from negative instance set if they do not  
%inv the negative sampling strategies to generate an appropriate set of negative instances to well train node-level embeddings under the setting of unsupervised contrastive learning. %we design CSGCN (Adaptive sampling graph contrastive network), an unsupervised graph representation learning framework. 
Particularly, we respectively resolve the class collision issue and the imbalanced negative data distribution issue. Extensive experiments are performed on three real-world datasets and the proposed approach achieves the SOTA model performance.

\end{abstract}

\section{Introduction}
%图神经网络

% graph embedding important\\
% better unsupervised, label cost, no label\\
% contrastive learning, a natural choice,\\
% graph data very few. \\
% illustrate two related work.
% shortcomings:
% 1. image, distance metrics
% 2. ring:shortcoming: not graph, image
% basics:
% 1. random walk generate sub-graphs 
% issue
% node2vec, graph2vec, etc, for unsupervised;
% differently, mutual information based.
% motivation:
% 1. 3 papers on graph negative sampling strategy

% issues:
% 1. node embeddings close to graph embeddings;  X

% should be 
% node embedding far away from negative embeddings;
% however, should under a generative model contraints. 

% -> under generative model: positive node close to each other;
% negative nodes far away from . 

% how to:
% 1. determine possible positive samples from negative samples to avoid class collision. 
% positive distribution
% negative distribution
% 2. two level dpp to evenly sample the negative space. 
% 3. to further measure the importance of negative samples, weighted...
% negative: class collision: reduce loss
% pseudo 
% issue: multiple negative sampling
% 

In the literature, various graph neural network (GNN) models have been proposed for graph analysis tasks, such as node classiﬁcation \cite{2017GCN}, link prediction \cite{2018link} and graph classification \cite{2019structpool}. 
%Graph data analysis has long been investigated with ﬂourishing results \cite{2016node2vec}\cite{2017chebyshev}. Recently, with the rapid development of deep learning, various graph neural network (GNN) models have been proposed for graph analysis  related research issues, such as node classiﬁcation \cite{2017GCN}, link prediction \cite{2018link} and graph classification \cite{2019structpool}. 
Generally, most existing GNN-based approaches are proposed to train, in a supervised manner, graph encoder to embed localized neighboring nodes or node attributes for a graph node into the low-dimensional feature space. %targeting at learning the powerful
By convoluting $K$-hops neighboring nodes, adjacent nodes naturally have similar feature representations.
% as they are ``geometrically'' close to each other although not under the Euclidean space. 
Notably, the consequent downstream tasks inevitably rely on the quality of the learnt node embeddings. 

% Graph data analysis has long been investigated with ﬂourishing results \cite{2016node2vec}\cite{2017chebyshev}. Recently, with the rapid development of deep learning, various graph neural network (GNN) models have been proposed for graph related research issues, such as node classiﬁcation \cite{2017GCN}, link prediction \cite{2018link} and graph classification \cite{2019structpool}. Generally, GNN based approaches try to train a transductive or an inductive graph encoder to learn low-dimensional representations of graph nodes by considering either graph structural information or node attributes. The corresponding graph convolution operations are then performed to aggregate embeddings of neighboring nodes. %, and aggregate these feature embeddings. 
% By doing so, neighboring graph nodes naturally have similar feature representations as they are ``geometrically'' close to each other, %as their information are convoluted with each other. 
% and various consequent downstream tasks, e.g., node classiﬁcation, could be performed on top of the learnt feature embeddings. Apparently, the model performance largely relies on the quality of the learnt feature embeddings.

%无监督为什么重要
For many real graph applications, e.g., protein analysis \cite{2017ppi}, it intuitively requires an unavoidable cost or even the specialized domain knowledge to manually annotate sufficient data to well train the graph encoders by minimizing the loss associated with the specified supervised learning task. Alternatively, a number of milestone unsupervised random walk based GNNs, including but not limited to node2vec \cite{2016node2vec} and graph2vec \cite{2016subgraph2vec}, are consequently proposed towards training the universal node embeddings and then various supervised downstream tasks are directly applied on these node embeddings. %top of the learnt unsupervised node embeddings. 
% Generally, GNNs require sufficient labeled graph data to train the graph encoder for generating discriminative node embeddings. However, these annotations vary in difficulty and cost and even require specialized domain knowledge, such as \blue{biology, chemistry and sociology}. To alleviate this issue, several attempts have been made towards training GNNs in an unsupervised manner. Currently, the unsupervised method applied on graph data use the loss function based on adjacency relationship prediction or structure reconstruction so that neighboring nodes have similar node representations. However, these tasks have proved to have limited effects \cite{2018DGI}. 
%以往的无监督方法，提到对比学习，我们比他们好在哪里
Similarly, another line of unsupervised graph representation learning approaches, i.e., graph kernel based methods, also utilizes the graph structural information to embed graph nodes with similar structures into similar representations. 
Most recently, the contrasive learning \cite{2020momentum,2020hard} is originally proposed to learn feature embeddings for each image in a self-supervised manner. To this end, these proposed approaches first generate two random augmentations for the same image and define these two as a pair of positive samples, and simply treat samples augmented from other images as negative samples.
% To this end, these proposed approaches first generate positive instances by random original images (e.g., by rotation), and %and negative instances respectively and simply treat images other than itself as negative instances.
Then, the contrastive loss is designed to maximize the mutual information between each pair of positive samples. %the input and the generated positive instances; meanwhile to minimize the mutual information between the input and the negative instances.
The learnt embeddings are believed to well preserve its inherent discriminative features. Research attempts are then made to adapt the successful contrastive learning approaches to unsupervised graph representation learning problem \cite{2020multi-view,2018DGI}. In \cite{2018DGI}, the graph-level representation is built to contrast with each node representation to acquire node representations fitting for diverse downstream tasks.
%and this allows the learning of a more robust node-level representation fitting for diverse downstream tasks.
\cite{2020multi-view} adopts diffusion graph as another view of the given graph, and the contrast is performed between graph representation of one view and node representation of another view.
%\cite{2020multi-view} constructs the view-level graph representation, and the contrast is performed between view-level graph representation and node representation. However, this might drive the learnt node embedding to be close to each other %to some extent 
As all the node embeddings are forced to approximate the same graph representation.  Intuitively, a coarser level graph analysis task, e.g., graph classification, would benefit a lot from such kind of contrastive learning, whereas a fine-grained level task, e.g., node classification, might not benefit that large. % \orange{because of the loss of discrimination}. 

% Recent advances in contrastive learning have given rise to human-level performance in several computer visual (CV) \cite{2020momentum} or natural language process (NLP) \cite{2013NLP} tasks under unsupervised scenario. However, there is few literature \cite{2020multi-view} focusing on contrastive learning applied on graph data. In these approaches, \cite{2018DGI}\cite{2020multi-view} trains a GNN model to maximize the mutual information between a high-level ``global'' graph representation and ``local'' node embeddings. This encourages the graph encoder to carry the type of features or structure pattern that appearing in any subgraphs. Such methods are very intuitive for the effectiveness of the graph-level task, because these patterns that appear in any subgraphs of graph are the most significant feature that distinguish graphs from each other. However, for node representation, based on method that make all nodes ``close'' to the same graph representation, all node representation ``close'' to each other also, losing the uniqueness of the node representation, which is very important on the node-level tasks.

%结论
To address aforementioned research gap, this work is thus motivated to investigate whether the contrastive learning could be effectively carried on in a node-wised manner. %That is,\
That is, for each graph node $x$ to be embedded, our desired contrastive learning is to maximize the mutual information between $x$ and its positive examples $x^+$ instead of a graph representation, %(random augmentations of $x$), 
and simultaneously to minimize the mutual information between $x$ and its negative examples $x^-$. %However, \
Meanwhile, there exist two research challenges to be addressed. First, the sampled negative examples $x^-$ might contain the true positive example $x^+$ which is known as class collision issue. Second, how the density of negative samples will affect the contrastive learning has not been studied.
% how to appropriately sample the negative examples such that a dedicate hyper plane of class boundary could be naturally sensed by the employed GNN models is seldom investigated in the literature. %To address aforementioned issues, Although $x^+$ is usually generated from the random augmentations from the set of $K$-hop neighboring nodes of $x$, 
We assume that the underlying true positive examples could be statistically similar to $x$, i.e., unseen positive examples should obey the same prior probability distribution as $x$. Similarly, the multiple typed negative examples $x^-$ are assumed to obey different probability distributions. With this assumption, the class collision issue could be intuitively resolved by removing those examples from the set of $x^-$ they are more likely generated by the assumed positive data distribution. For the second point, it can be known from the contrasitive loss that $x$ will be farther away from feature area with dense $x^-$ than area with sparse $x^-$. The density distribution of $x^-$ is used as factor to determine the distance of $x$ and $x^-$ is questionable. Therefore, after removing negative examples in doubt, a subset of negative examples should be diversely sampled to form $x^-$ for the contrastive learning. Thus, this paper proposed an adaptive negative sampling strategy for the learning of the node embedding in a node-wised contrastive learning manner.
%\orange{Therefore, we proposed an node-level adaptive negative sampling strategy to eliminate possible positive samples from $N$ negative samples. The key to our sampling strategy is to directly fit the distribution of positive and negative examples, and use the ratio of the two probabilities to eliminate negative samples.}
% That is, for each graph node $x$ to be embedded, our desired contrastive learning is to maximize the mutual information between $x$ and its positive examples $x^+$, which usually are its augmentations of , and assign $N$ negative examples low similarity when true labels of the downstream task are typically not available. Such a way of dividing positive and negative samples will greatly damage the representation quality since $x^-$ may be actually similar to $x$. Therefore, we proposed an node-level adaptive negative sampling strategy to eliminate possible positive samples from $N$ negative samples. The key to our sampling strategy is to directly fit the distribution of positive and negative examples, and use the ratio of the two probabilities to eliminate negative samples.
The major contribution of this paper are summarized as follows.
\vspace{-0.1cm}
\begin{itemize}
\item  To the best of our knowledge, this paper is among the first attempts to propose a node-wise contrastive learning approach to learn node embedding in an unsupervised manner. In the proposed approach, positive samples and negative samples are assumed to obey different data distributions, and the class collision issue could be addressed by eliminating in doubt negative samples if they are more likely generated by a positive data distribution. 
\item We propose a determinantal point process based negative instances sampling strategy which is believed to be able to sample diverse negative examples to be contrasted to learn node embedding. 
% Meanwhile, we theoretically analyze the loss bound of our approach. 
% \item  We exhaustively ablate our approach and show the generalization and utilization of the learnt representations for a wide range of hyper-parameters.

\item We perform extensive experiments on several benchmark datasets and the promising results have demonstrated that the proposed approach is superior to both baseline and the state-of-the-art approaches.
\end{itemize}

The rest of this paper is organized as follows. Section \ref{sec:related} reviews related work and then we formulate the problem in Section \ref{sec:pre}. The proposed approach is detailed in Section \ref{sec:approach}. Experimental results are reported in Section \ref{sec:exp} and we conclude the paper in Section \ref{sec:conc}.

% \vspace{-0.4cm}
\section{Related Work}
\label{sec:related}
% We categorize related work into graph representation learning, contrastive learning and negative sampling which are briefly reviewed in the following subsections. 
% \vspace{-0.2cm}
\subsection{Graph Representation Learning}
\subsubsection{Supervised Methods} 

The earlier graph representation learning attempts have been made in the supervised settings. ChebyNet \cite{2017chebyshev} and GraphWave \cite{2020heat_kernel} leverage graph Fourier transformations to convert graph signals %defined in vertex domain 
into spectral domain. Kipf and Welling \cite{2017GCN} propose graph convolutional network (GCN) via a localized first-order approximation to ChebyNet \cite{2017chebyshev}, and extend %convert the meaning of 
graph convolution operations to the aggregation of neighbor nodes. To further the success of GCN, GAT \cite{2018gat} and GeniePath \cite{2018geniepath} are proposed to sample more informative neighbor nodes for the convolutions. There also exist some approaches targeting at resolving efficiency issue  \cite{2018graphsage,2018fastgcn}. %have been proposed to increase the computing speed or reduce the memory usage.
\subsubsection{Unsupervised Methods}
The unsupervised graph representation learning methods could be classified into \textit{random walk-based} methods \cite{2015Line,2014deepwalk} and \textit{graph kernel-based} methods \cite{shortest_kernel,graphlet_kernel}. The random walk-based methods are applied for each graph node and the nodes in a sequence of walk are to be encoded. By dosing so, the neighboring nodes generally are trained to have similar embeddings regardless of the graph structural information as well as the node attributes. Such kinds of methods are usually transductively performed and thus need to be re-trained to represent the unseen nodes which inevitably limits their wide applicability. %possibilities to be applied in more applications.  %Meanwhile, to obtain the representation of new nodes have not appeared in training stage, the entire network must be retrained, which limit the scalability of model. Graph Kernel methods decompose graphs into well-designed substructures and use kernel function to measure graph similarity between them. Nevertheless, the design of these substructure requires a full understanding and professional knowledge of graph.

\subsection{Contrastive Learning and Negative Sampling}
Contrastive learning is recently proposed to learn feature embeddings in a self-supervised manner. %  selfa natural choice to capture the similarity from data, which rely on a number of pre-defined positive samples and negative samples for generating good quality representations. 
The quality of the learnt embeddings largely replies on the generated positive instance set and the negative instance set. Accordingly, various approaches have been proposed with a focus on constructing positive samples. In the domain of NLP, %Therefore, the main difference between methods are the way to define positive pairs. In the language domain, 
\cite{2018sentence} treats the contextual sentences as positive pairs. For the domain of image recognition, a good number of research works are proposed to train encoders to discriminate positive samples from negative samples. For graph data, \cite{2016node2vec} encodes graph nodes using a generalized proximity %in the graph, 
and treats these nodes not appearing in $w$ steps of a random walk or without common neighbors as negative examples. In \cite{2017struc2vec}, two nodes that have a similar local structure are considered as positive sample pairs despite the node position and its attributes. 
%The success of contrastive learning naturally calls for research attentions to investigate the negative sampling strategies which are briefly reviewed in the following subsection.  
%many tasks are based on the idea of instance discrimination, which is proved to be linked to mutual information, such as position prediction, color prediction, rotation prediction and many other ``pretext'' objectives. The process of training the encoder to distinguish positive samples from negative samples is to make the mutual information with the positive samples increase, and the mutual information with the negative samples reduce. 
Then, several SOTA approaches have been proposed to adapt contrastive learning on graph data \cite{2018DGI,sun2020infograph,2020multi-view,2020GCC}. DGI \cite{2018DGI} maximizes the mutual information between the node embeddings and graph embeddings. %, and InfoGraph \cite{sun2020infograph} proposes a semi-supervised contrastive learning method. % for extends DGI to graph-level representation and semi-supervised scenario.
% A pre-trained GCC \cite{2020GCC} using contrastive learning has demonstrated satisfactory performance in various down stream tasks. %This key point behind the good performance of the contrastive learning model is how to define positive and negative samples in the case of an unlabeled datasets, so that the goal of the training stage is consistent with downstream tasks.

%\subsection{Negative Sampling on Graph}\cite{2014deepwalk}
%To investigate how to further enhance the quality of the learned embeddings, 
InfoGraph %\cite{2018DGI}
\cite{sun2020infograph} treats nodes that are virtually generated by shuffling feature matrix %or corrupted adjacency matrix 
as negative samples. Mvgrl \cite{2020multi-view} further defines two views on graph data and graph encoders are trained to maximize mutual information between node representations of one view and graph representations of another view and vice versa. \cite{2020GCC} consider two subgraphs augmented from the same $r$-ego network as a positive instance pair and these subgraphs from different $r$-ego as negative sample pair, where $r$-ego represents the induced subgraph containing the set of neighbor nodes of a given node $v$ within $r$ steps.

\section{Preliminaries and Problem Formulation}
\label{sec:pre}
%In this section, we first illustrate the contrastive learning setup graph data and then briefly review preliminary background knowledge about the Determinantal Point Process (DPP) in the following subsections.
In this section, we first briefly review the Determinantal Point Process (DPP) \cite{dpp} adopted to diversely sample negative instances, then we describe the notations as well as the problem setup. 

\subsection{Determinantal Point Process}
The original DPP is to proposed to model negatively correlated random variables, and then it is widely adopted to sample a subset of data where each datum in this set is required to be correlated with the specified task, and simultaneously is far away from each other. Formally, let $\mathcal{P}$ denote a probability distribution defined on a power set $2^Y$ of a discrete ﬁnite point set $Y$ = \{$1,2,...,M$\}. $Y\sim\mathcal{P}$ is a subset composed of data items randomly generated from $\mathcal{P}$. Let $A$ be a subset of $Y$ and $B \in \mathbb{R}^{M\times M}$ be a real positive semi-definite similarity matrix, then we have

%-----------------------
\begin{equation}
\mathcal{P}(A \subseteq Y) = |B_A|,
\end{equation}
%-----------------------

where $B_A$ is a sub-matrix of $B$ indexed by the elements of subset $A$. $|\cdot|$ denotes the determinant operator. If $A = \{i\}$, $\mathcal{P}(A \subseteq Y)=B_{i,i} $; and if $A = \{i,j\}$, $\mathcal{P}(A \in Y)$ can be written as 
%-----------------------
\begin{equation}
\mathcal{P}(A \subseteq Y) = 
\begin{vmatrix} B_{i,i} & B_{i,j}\\ B_{j,i} & B_{j,j} \end{vmatrix} = 
\mathcal{P}(i \in Y)\mathcal{P}(j \in Y)-B_{i,j}^2,
\end{equation}
%-----------------------

Thus, the non-diagonal matrix entries represent the correlation between a pair of data items. The larger the value of $B_{i,j}$, the less likely that $i$ and $j$ appear at the same time. Accordingly, the diversity of entries in the subset $A$ could be calculated. As for our approach, DPP is adapted to sample a evenly distributed subset from negative instance set. %, and the distribution of node embeddings in the subset is scattered.
\subsection{Notations and Definitions}%Notation and Definition}
Let $G = (V,E)$ denote a graph, $V$ denote the node set containing $N$ nodes, $E \subseteq V\times V$ denote the edge set where $e=(v_i, v_j) \in E$ denotes an edge between two graph nodes, $\mathrm{X} = \{x_1,...,x_d\}$ denote the node feature set where $x_i \in \mathbb{R}^{d_{in}}$ represents the features of node $v_i$. The adjacency matrix is denoted as $\mathrm{A}\in \mathbb{R}^{N \times N}$, where $\mathrm{A}_{ij} = 1$ represents that there is an edge between $v_i$ and $v_j$ in the graph and 0 otherwise. 
%And the node features set is $\mathrm{X} = \{x_1,...,x_N\}$, where $N$ is the number of nodes in the graph and $x_i \in \mathbb{R}^{d_{in}}$ represents the features of node $v_i$. 
%The relational information are provided by the adjacency matrix $\mathrm{A}\in \mathbb{R}^{N \times N}$, where $\mathrm{A}_{ij} = 1$ represent there is an edge $i\to j$ in the graph and $\mathrm{A}_{ij} = 0$ otherwise. 
For a given node $v_i$, its $K$-hops neighbor set is denoted as $\mathcal{N}_K(v_i)$ containing all neighboring nodes of $v_i$ within $K$ hops, defined as $\mathcal{N}_K = \{v_j:d(v_i,v_j)\leq K\}$ where $d(v_i,v_j)$ is the shortest path distance between $v_i$ and $v_j$ in the graph $G$. Then, %Based on the neighbor set $\mathcal{N}_K(v_i)$, 
the induced subgraph is defined as follows. 
%-----------------------
\begin{definition}
\textbf{Induced subgraph $s$}. Given $G=(V,E)$, a subgraph $s = (V^{\prime}, E^{\prime})$ of $G$ %is a pair of $(V^{\prime}, E^{\prime})$ where $V^{\prime} \subseteq V$ and $E^{\prime} \subseteq E$. 
is said to be an induced subgraph of $G$ if all the edges between the vertices in $V^{\prime}$ belong to $E^{\prime}$. 
\end{definition}
\subsection{Problem Setup}
Given a $G$, our goal is to train a graph encoder $\mathcal{G}:\mathbb{R}^{N\times F} \times \mathbb{R}^{N\times N} \to \mathbb{R}^{N\times d}$, such that $\mathcal{G}(\mathrm{X},\mathrm{A}) = \mathrm{H}= \{h_1,...,h_N\}$ represents the low-dimensional feature representations, where $h_i$ denotes the embeddings of $v_i$. Then, the learnt $\mathcal{G}$ is used to generate node embeddings for downstream tasks, e.g. node classification.  

The purpose of our approach is to maximize the mutual information between a pair of positive instances, and minimize the mutual information between a pair of negative and positive instances simultaneously. 
% Since our goal is to increase the mutual information of positive sample pair, and minimize the mutual information of negative sample pair, the implicit meaning behind this loss function is to enable the encoder to capture the latent low-dimensional feature consistent in positive examples.
Similar to the infoNCE \cite{2019CPC}, the general form of our unsupervised contrastive learning is to minimize the contrastive loss, given as
%Because the model is trained under the condition of unsupervised comparative learning, we cannot use the loss function widely used in graph semi-supervised learning. Inspired by recent success of contrastive learning in CV and NLP, we use the contrastive loss infoNCE \cite{2019CPC} to optimize our model.
%-----------------------
\begin{equation}\label{con:loss_original}
{L}= -\sum_{i=1}^N \log \frac{e^{f({h}_i^q,{h}_i^k)/\tau}}{ e^{f({h}_i^q,{h}_i^k)/\tau}+\sum_{j\not=i}^N e^{f({h}_i^q,{h}_j^k)/\tau}},
\end{equation}
%-----------------------
where $f(\cdot,\cdot)$ is the score function to score the agreement of two feature embeddings, %the embeddings between nodes.
and in our approach the score function is simply the dot product calculated as $f({h}_i^q,{h}_j^k) = h_i^q\cdot {h_j^k}^T$, and $\tau$ is the temperature hyper-parameter. $h_i^k$ is a positive instance of $h_i^q$ , and two of them are usually defined as two different random augmentations of the same data. In our model, we define $h_i^q,h_i^k$ are two random augmented embeddings of $v_i$. Note that the first term of the denominator is to maximize the mutual information between a pair of positive instances, and the second term is to minimize the mutual information between a pair of negative and positive instances. By optimizing Eq. \ref{con:loss_original}, the employed model is believed to be able to learn the most discriminative features that are invariant to positive instances.

Nevertheless, there are some problems to be address. Intuitively, we hope that two nodes that have the same label in the downstream task should have similar embeddings. However, we treat $h_{j \not= i}^k$ as negative sample and try to be away from it in feature space while $v_j$ may have the same label as $v_i$, which is called class collision. At the same time, we noticed that the current negative sampling strategy ignores the influence of the density of embedding distribution of negative samples. A node embedding will be updated to be farther away from feature subspace where its negative nodes are more densely distributed. Therefore, we designed approach to adaptive sampling negative samples to avoid those problems described above.

\section{The Proposed Approach}
\label{sec:approach}

\subsection{Graph Embeddings}\label{sec:generation}
The proposed node-wised contrastive learning scheme allows various choices of the graph neural network architectures. We opt for simplicity reason and adopt the commonly adopted graph convolution network (GCN) \cite{2017GCN} as our graph encoder $\mathcal{G}$. 
\subsubsection{Augmentation}
By following \cite{2020GCC}, we first employ a $k$-steps random walk on $G$ starting from a specific node $v_i$, and a sequence of walking nodes $seq_i$ = \{$t_1,...,t_k$\} is used to form the set of vertices $V^{\prime}$. The subgraph $s_i$ induced by $V^{\prime}$ is regarded as a random augmentation of node $v_i$. Then, we repeat aforementioned procedure and eventually we generate two induced subgraphs $s_i^q, s_i^k$, those embeddings are respectively denoted as $h_i^q$ and $h_i^k$ and regarded as a positive pair.

% as the positive instances of $v_i$, and their embeddings are respectively denoted as $h_i^q$ and $h_i^k$.

\subsubsection{Encoder}
%Our framework allows various choices of the network architecture without any constraints. 
The employed GCN layers are defined as  $\sigma(\widetilde{\mathrm{A}}_{i}\mathrm{X}_{i}W)$ which is used to embed node $v_i$, where $\widetilde{\mathrm{A}}_i = {\widehat{\mathrm{D}}_{i}}^{-\frac{1}{2}}  \widehat{\mathrm{A}}_{i}^{}{\widehat{\mathrm{D}}_{i}}^{-\frac{1}{2}} \in \mathbb{R}^{n_i\times n_i }$ is symmetrically normalized adjacency matrix of a subgraph $s_i$. $\widehat{\mathrm{D}}_{i}$ is the degree matrix of $\widehat{\mathrm{A}}_{i} = {\mathrm{A}}_{i}+\mathrm{I}_{n_i}$, where ${\mathrm{A}}_{i}$ is the original adjacency matrix of $s_i$, %$\mathrm{I}_{n_i}$ is the identity matrix. 
$\mathrm{X}_{i} \in \mathbb{R}^{n_i\times d}$ is the initial features of nodes in $s_i$, $W \in \mathbb{R}^{d_{in} \times d}$ is network parameters, $\sigma$ is a ReLU \cite{2011relu} non-linearity and $n_i$ is the number of nodes contained in $s_i$. Putting $\widetilde{\mathrm{A}}_{i}, \mathrm{X}_{i}$ into graph layer to perform convolution operation and then we could acquire node embeddings $H_i \in \mathbb{R}^{n_i\times d}$ of subgraph $s_i$.

\subsubsection{Readout}
After convolution operation of GNN layers, we feed the embedding set $H_i$ into the readout function $\mathcal{R}(\cdot)$ to compute an embedding of $v_i$. The readout function adopted in the experiments is given as follows
%-----------------------
\begin{equation}
\mathcal{R}(H_i)= \sigma(\frac{1}{n_i}\sum_{j=1}^{n_i}h_{i,j}+\max(H_i)),
\end{equation}
%-----------------------
where $h_{i,j}$ represents the $j$-th node embedding in $H_i$, $\max(\cdot)$ simply takes the largest vector along the row-wise and $\sigma$ is the non-linear sigmoid function. %Such a simple readout function has been proved in experiments to achieve good results on most datasets. 
Eventually, the node embedding is acquired as $h_i = \mathcal{R}(H_i)$. The node encoding process is illustrated in $\text{Algorithm}$ \ref{algo:augmented process}.

%-----------------------
\begin{algorithm}[thb]
\label{algo:augmented process}
\caption{Generate embeddings of the augmented instances}
\KwIn{graph $G$, adjacency matrix $\mathrm{A}$, feature matrix $\mathrm{X}$,\\
\setlength{\parindent}{3em} graph encoder $\mathcal{G}$, score function $f$, readout function $\mathcal{R}$,\\
\setlength{\parindent}{3em} random-walk operator $\mathcal{RW}$, subgraph sampler $\Gamma_{sub}$, concatenator $\parallel$
}
\KwOut{The node embeddings $\mathrm{H}^q \in \mathbb{R}^{N \times d}, \mathrm{H}^k \in \mathbb{R}^{N \times d}$}
\SetAlgoLined
 Initialize $\mathrm{A}$ and $\mathrm{X}$\;

\For{ i=1 to $N$}
{
    $s_i^q$ = $\mathcal{RW}(v_i)$, $s_i^k$ = $\mathcal{RW}(v_i)$     \\
    $\mathrm{A}_i^q, \mathrm{X}_i^q$ = $\Gamma_{sub}(s_i^q,\mathrm{A},\mathrm{X})$\\
    $\mathrm{A}_i^k, \mathrm{X}_i^k$ = $\Gamma_{sub}(s_i^k,\mathrm{A},\mathrm{X})$\\ 
    $H_i^q$ = $\mathcal{G}(\mathrm{A}_i^q, \mathrm{X}_i^q)$, $H_i^k = \mathcal{G}(\mathrm{A}_i^k, \mathrm{X}_i^k)$\\
    $h_i^q = \mathcal{R}(H_i^q)$, $h_i^k = \mathcal{R}(H_i^k)$\\
}
\textbf{return} $\mathrm{H}^q$ = $\mathop{\big\Vert}\limits_{i=1}^{N}h_i^q$, $\mathrm{H}^k$ = $\mathop{\big\Vert}\limits_{i=1}^{N}h_i^k$\\
\end{algorithm}
\begin{figure}[tb]
\setlength{\abovecaptionskip}{5pt}
\setlength{\belowcaptionskip}{0pt}
\centering
\subfigure[Initial stage: target node $v$ (in red). %The division of positive and negative samples before adding positive samples.Left: a target node $v$ (in red) to embed and the rest are negative nodes (in blue); Right: the feature embedding space.
%  The division of positive and negative samples before adding positive samples
] {
\label{fig:a}
\includegraphics[width=2.in]{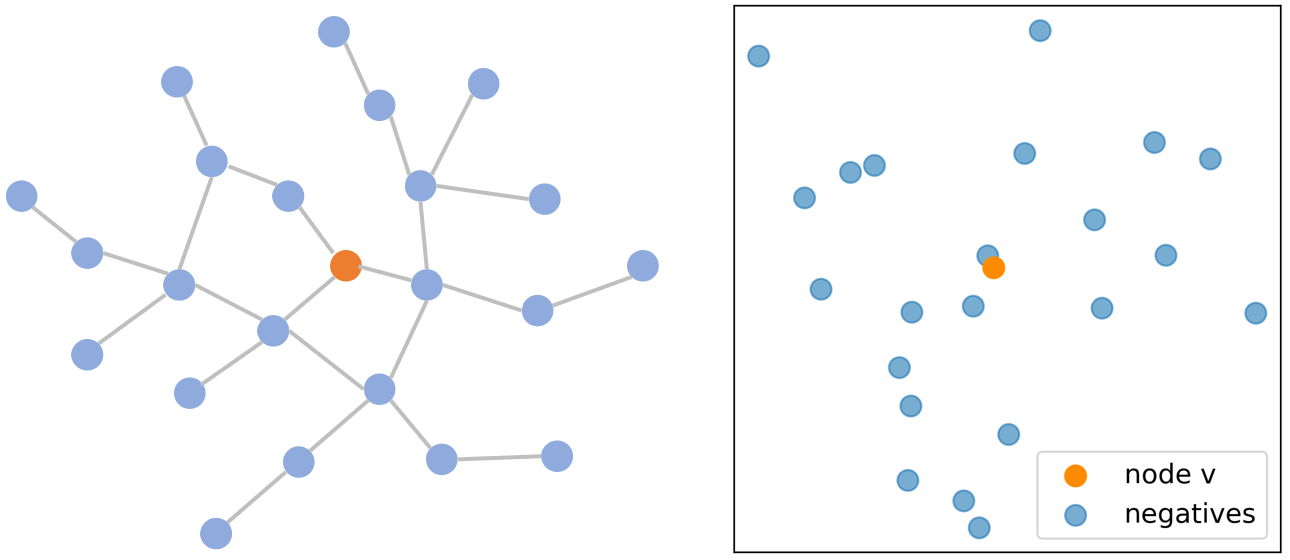}}
\quad
\subfigure[The augmented positive instances. %The division of positive and negative samples after adding neighboring and mixup positive samples
]{
\label{fig:b}
\includegraphics[width=2in]{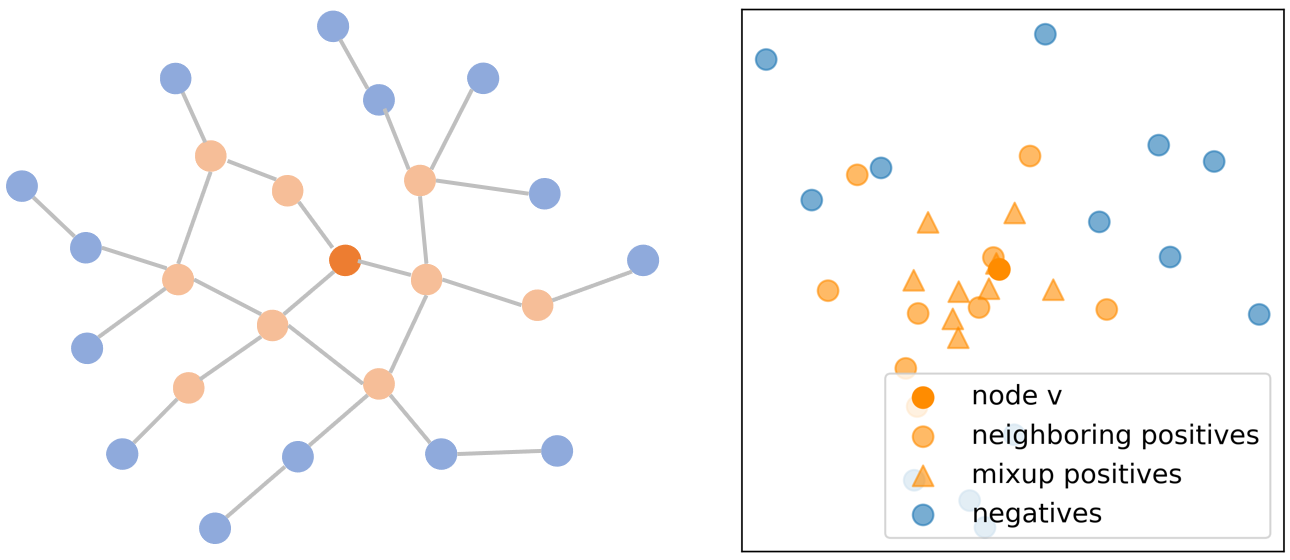}}
\quad
\subfigure[Fit the embedding distribution for positive and negative samples.]{
\label{fig:c}
\includegraphics[width=2in]{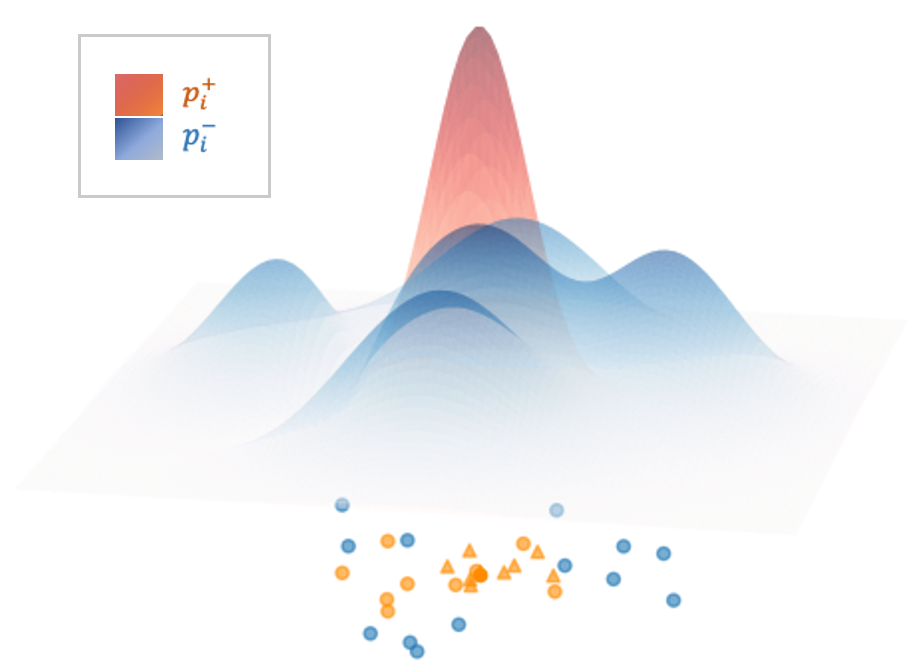}}
\quad
\subfigure[Resample the negative samples. %Resample negative samples
]{
\label{fig:d}
\includegraphics[width=2in]{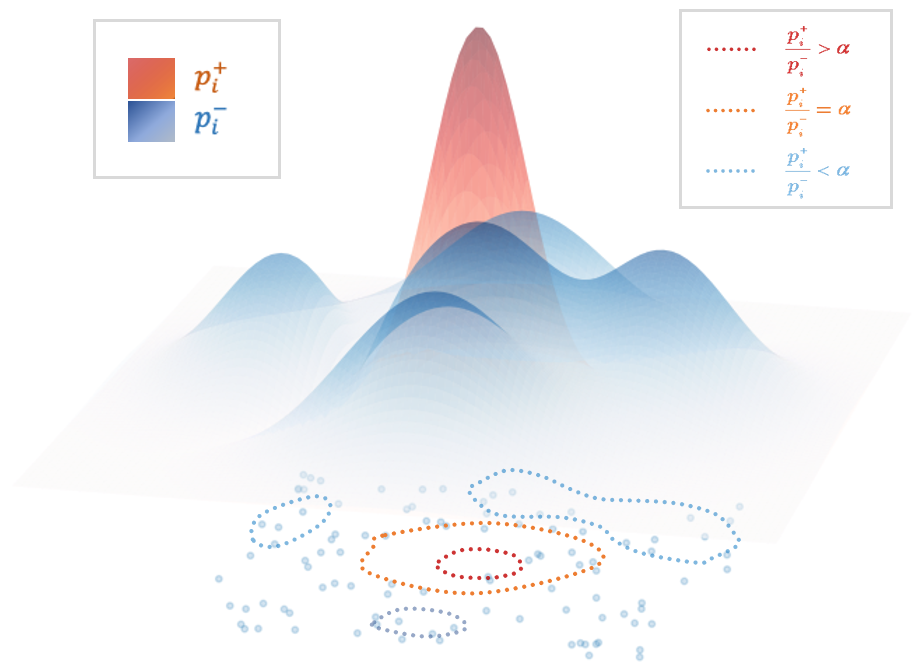}}
\caption{Process of generating positive and negative samples by our approach. In each subfigure, the left depicts the graph topological structure and the right plots the feature embedding space where each colored dot represent the embeddings of a positive or negative data node. In the initial stage, plotted in \textbf{(a)}, there is only one target node treated as positive node (in red) and the rest are negative nodes (in blue). In \textbf{(b)}, the positive nodes are augmented by adding ``in-doubt'' nodes (in orange dot) and mixup positive nodes (in orange triangle), and from the right figure, it is noticed that the embeddings of orange dots and triangles are close to that of the target node (red dot). And orange and blue nodes could be well separated which indicates two distributions could be estimated. In \textbf{(c)}, the underlying positive instance distribution and negative instance distribution could be well fit using these data. In \textbf{(d)}, the dashed loop is the contour of $\frac{p_i^+(v_j)}{p_i^-(v_j)}$. Note that the smaller the orange dashed loop, the more confident that datum falling a positive instance. 
%and \textbf{(b)}, the left side is the partial adjacency information of target node $v_i$ and the right side is the embeddings distribution for $v_i$ and its neighbors where the orange ones represent the positive samples and the blue ones represent the negative samples. In \textbf{(d)}, the dashed line is the contour of $\frac{p_i^+(v_j)}{p_i^-(v_j)}$.
}
\label{fig:negative_sampling}
\end{figure}
% -----------------

\subsection{Resolving Class Collision}
\label{subsection:distribution}
%\red{discover possible positive instances from negative samples.\\1. generate negative examples;\\2. find possible positive.}
Given a node $v_i$, its negative and positive sample set are respectively denoted as $S_i^- = \{v_1,...,v_{i-1},v_{i+1},v_N\}$ and $S_i^+ = \{v_i\}$. To alleviate the class collision issue, it is desired to discover those ``in doubt'' negative samples that are more likely to belong to the same class of $v_i$. The overall procedure is depicted as below.  

%$v_j$ that of the same class as $v_i$ from the negative sample set. We first describe the flow of the algorithm, and then introduce the details of each part.
% Without loss of generality, we take $v_i$ as an example. At the beginning, we set its negative samples set as $S_i^- = \{v_1,...,v_{i-1},v_{i+1},v_N\}$ and positive sample set $S_i^+ = \{v_i\}$. In order to alleviate the situation of class collision, we need to remove some negative samples $v_j$ that of the same class as $v_i$ from the negative sample set. We first describe the flow of the algorithm, and then introduce the details of each part.
First, we assume that $S_i^+$ and $S_i^-$ respectively obey different prior probability distributions. The ``in-doubt'' negative examples are discovered if they are more likely to be generated by the data distribution of positive instances. To fit the embedding distribution $p_i^+$ and $p_i^-$, we employ two independent neural networks, i.e., $\mathcal{F}_i^+$ and $\mathcal{F}_i^-$ to fit the distributions. If the probability that $v_j$ belongs to $S_i^+$ is higher than the probability of being a negative instance, i.e., $\frac{p_i^+(v_j)}{p_i^-(v_j)}> \alpha$, we remove $v_j$ from $S_i^-$, where $\alpha$ is the soft-margin to discriminate an instance. Detailed steps are illustrated in the following paragraphs. %etermine control the strictness of removing negative samples. 
%$p_i^+$, $p_i^-$ of its positive samples and negative samples respectively. When the probability of $v_j$ being a positive sample is higher than the probability of being a negative sample to a certain extent, i.e., $\frac{p_i^+(v_j)}{p_i^-(v_j)}> \alpha$, we remove node $v_j$ from $S_i^-$, where $\alpha$ is a hyper-parameter to control the strictness of removing negative samples. 
\subsubsection{Forming the sample set $S_i^+$ and $S_i^-$}
Initially, the positive instance of a given node $v_i$ is also augmented by $v_i$ plotted in orange and the rest nodes plotted in light-blue are considered as negative instances, which is shown in Figure \ref{fig:a}. Apparently, not all the blue data are the true negative instances. To consider the unsupervised settings, it is acceptable to assume that the 
% closet 
``closest'' node to $v_i$ should have the same underlying class label. Therefore, a few nearest neighbor nodes $v_j\in \mathcal{N}_K$ are transited to $S_i^+$ from $S_i^-$. Using the mixup algorithm \cite{2018mixup}, more positive samples will be generated to further augment the positive instance set $S_i^+$.
% \red{A hyper-parameter is fine-tuned using the mixup algorithm \cite{2018mixup} to further augment the positive instance set $S_i^+$,}
and the results are illustrated in Figure \ref{fig:b}. 
%Meanwhile, we also adopted the mixup algorithm \cite{2018mixup} to $S_i^+$, further increasing the number of positive samples as Fig. \ref{fig:b}.

%Initially, we only see $v_i$ in graph as positive sample as Fig. \ref{fig:a}, which is difficult to fit a distribution. Therefore, we need to generate more positive samples to make the distribution of positive samples more accurate. Considering the graph structure is generally homogeneous, that is, the class of connected nodes are usually the same. So we add $v_j\in \mathcal{N}_K$ to $S_i^+$ and remove $v_j$ from $S_i^-$. Meanwhile, we also adopted the mixup algorithm \cite{2018mixup} to $S_i^+$, further increasing the number of positive samples as Fig. \ref{fig:b}.
\subsubsection{Fitting the positive and negative instance distribution} 
With the augmented $S_i^+$ and $S_i^-$, we employ two independent two neural networks $\mathcal{F}_i^+$, $\mathcal{F}_i^-$ to respectively fit the embeddings distribution for $v\in S_i^+$ and $v\in S_i^-$ as plotted in Figure \ref{fig:c}. To train $\mathcal{F}_i^+$, we treat it as a classifier, and data belong to $\mathcal{S}_i^+$ is assigned with a virtual label $class$ $1$, and data belong to $\mathcal{S}_i^-$ is virtually assigned with $class$ $0$. On the contrary, to train $\mathcal{F}_i^-$, we will assign label $class$ $1$ to the data belonging to $\mathcal{S}_i^-$ and assign label $class$ $0$ to the data belonging to $\mathcal{S}_i^+$. 

For a node $v_j\in S_i^-$, the output of $\mathcal{F}_i^+(h_j^k)$ and $\mathcal{F}_i^+(h_j^k)$ are respectively the probability that $v_j$ is a positive or negative instance of $v_i$. The ratio of these two probabilities with a soft-margin, calculated as $\frac{p_i^+(v_j)}{p_i^-(v_j)}> \alpha$, is adopted to determine whether $v_j$ should be removed from $S_i^-$ or not, and this soft-margin is plotted in the small orange dashed circle as shown in Figure \ref{fig:d}. 

\begin{figure}[htb]
\setlength{\abovecaptionskip}{5pt}
\setlength{\belowcaptionskip}{0pt}
\centering
\subfigure[Unevenly sampled negative instances. %Distorted distance between nodes
] {
\includegraphics[width=2in]{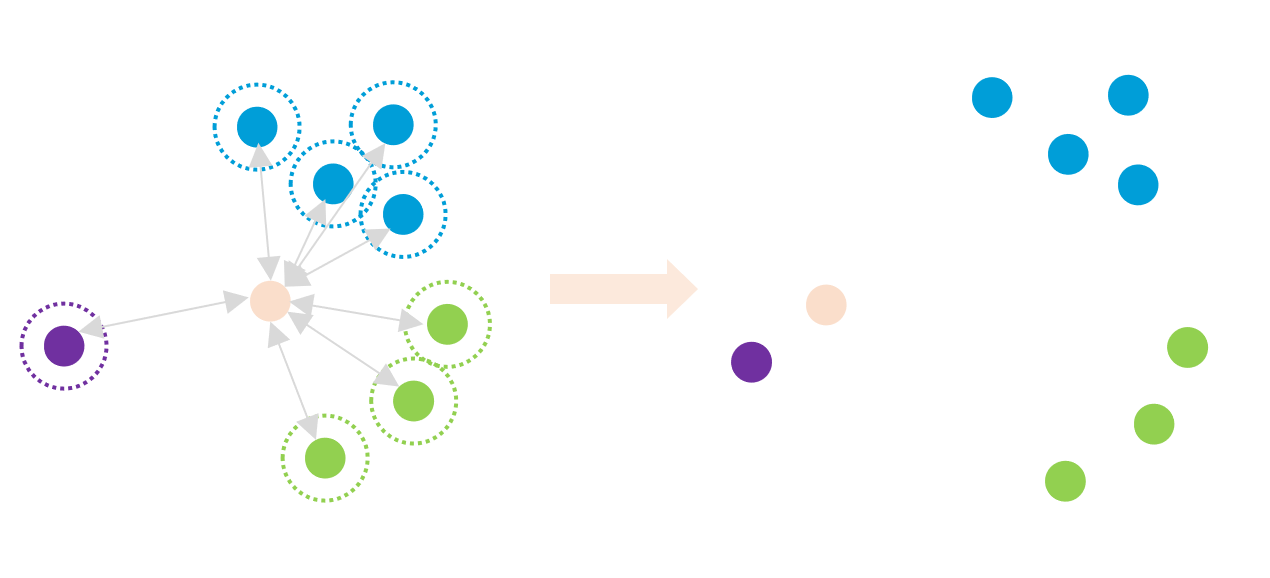}
% \subfigcapskip = 50pt
\label{fig:distored}}
\quad
\subfigure[Diversely sampled negative instances. %Reasonable distance between nodes
]{
\includegraphics[width=2in]{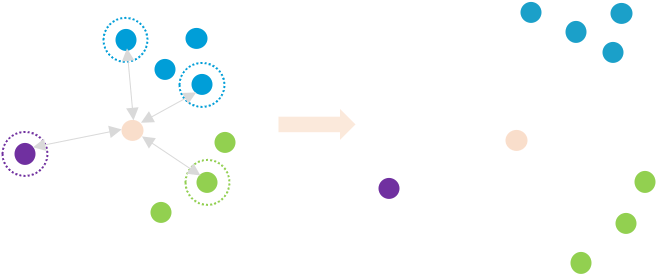}
\label{fig:dpp}}
\caption{The illustration of the effect of different sampling strategies. The dashed circle denotes that the corresponding node will be sampled. The pink colored dot is the embedding of target node $v$. \textbf{(a)} shows that for a current strategy, using all negatives, the learnt embeddings of $v$ will close to purple node whereas we desire that the embeddings of $v$ should, simultaneously, stay away from all negative instances as much as possible, as plotted in \textbf{(b)} where a diverse sampling strategy is applied, i.e. the proposed DPP strategy, on the embeddings space and reasonably ignores their distribution.
% underlying classes. %are the feature space before and after updated not using the DPP sampling algorithm, \textbf{(b)} are the feature space  before and after updated using the DPP sampling algorithm. The pink node is the query node and others are nodes in $S^-$. And those nodes surrounded by dashed lines are negative nodes that are sampled for training the embedding of query node. 
%Illustration of the results applied different contrastive strategy. \textbf{(a)} are the feature space before and after updated not using the DPP sampling algorithm, \textbf{(b)} are the feature space  before and after updated using the DPP sampling algorithm. The pink node is the query node and others are nodes in $S^-$. And those nodes surrounded by dashed lines are negative nodes that are sampled for training the embedding of query node. Refer to Section \ref{subsection:dpp_sampling} for more details.
}
\label{fig:contrastive strategy}
\end{figure}
% -----------------

%-----------------------
\begin{algorithm}[htb]
\label{algorithm2}
\caption{Generating positive instance set and negative instance set}

\KwIn{Adjacency matrix $\mathrm{A}$, node embeddings $\mathrm{H}^q$, $\mathrm{H}^k$,\\
\setlength{\parindent}{3em} hyper-parameter $K$, hyper-parameter $\alpha$,\\
\setlength{\parindent}{3em} $K$-hops neighbor node set \{$\mathcal{N}_K(v_1),...,\mathcal{N}_K(v_N)$\},\\
\setlength{\parindent}{3em} DPP sampler $\Gamma_{dpp}$, mixup operator $Mix$,\\
\setlength{\parindent}{3em} node embedding set $S$ = \{$h_1^k,...,h_N^k$\},\\
\setlength{\parindent}{3em} neural network set $\mathcal{F}^+ = \{\mathcal{F}_1^+,...,\mathcal{F}_N^+$\}, $\mathcal{F}^- = \{\mathcal{F}_1^-,...,\mathcal{F}_N^-$\}.\\
}
\KwOut{Positive samples sets \{$S_1^+,...,S_N^+$\},\\
\setlength{\parindent}{3.7em} negative samples sets \{$S_1^-,...,S_N^-$\}.\\
}
\SetAlgoLined
 initialization\;
 \For{i = 1 to $N$}
 {
    $S_i^-$ = $S \setminus \{h_i^k, i \in \mathcal{N}_K(v_i)\}$\\ 
    $S_i^+$ = $\{h_i^k, i \in \mathcal{N}_K(v_i)\}$\\
    $S_i^+$ = $Mix(S_i^+)$\\
 }
 \For{i = 1 to $N$}
 {
    $p_i^-$ = $\mathcal{F}_i^-(S_i^-)$\\
    $p_i^+$ = $\mathcal{F}_i^+(S_i^+)$\\
    \For{ j = 1 to $N$}
    {
        \If{$j \notin \mathcal{N}_K(v_i)$ and $\frac{p_i^+(v_j)}{p_i^-(v_j)}> \alpha$}
        {
            $S_i^-$ = $S_i^- \setminus \{h_j^k\}$,\\ 
            $S_i^+$ = $S_i^+ \cup \{h_j^k\}$\\
        }
    }
    $S_i^-$ = $\Gamma_{dpp}(S_i^-)$
 }
 \textbf{return} \{$S_1^+,...,S_N^+$\}, \{$S_1^-,...,S_N^-$\}
\end{algorithm}
% \vspace{-0.2cm}
%-----------------------
\subsection{Sampling Diverse Negative Examples}
\label{subsection:dpp_sampling}
As illustrated in Fig \ref{fig:contrastive strategy}, we can regard the process of contrastive learning as the interaction of forces between positive and negative samples. For the worst case of using all instances in $S_i^-$ for comparison, where the embedding distribution is seriously imbalanced as Fig \ref{fig:distored}, the updated $h_i$ will be farther away from the feature subspace where negative samples densely distributed. Intuitively, the comparison result between positive and negative samples should not be related to the density of negative samples. 
% the generated negative instance set $S_i^-$ is seriously imbalanced and consequently undermines the effect of the constrast learning performed between positive instances and negative instances. 
To cope with this distorted updata result, we adapt the Determinant point process (DPP) \cite{dpp} to our problem. In Fig. \ref{fig:dpp}, the DPP algorithm is applied to $S^-$ to sample a negative instances subset, where sampled negative instances spread across the entire feature space. In this case, the node embedding can avoid the influence of the density of feature space and be evenly away from each negative sample. We set the correlation between $h_i^q$ and each negative instance in $S_i^-$ is equally set to a constant. To calculate the similarity between negative instances, the Euclidean distance is adopted to  measure the pair-wise distance, computed as $d(h_i^q,h_j^k) = \sqrt{\sum_{l=1}^d ({h_{i,l}^q}-{h_{j,l}^k})^2}$.

\subsection{Node-wise Contrastive Learning Loss}
As pointed in \cite{2020hard}, different nodes contribute differently to the unsupervised contrastive learning. We are therefore inspired to further differentiate the importance of the diversely sampled negative instances to our node-wise contrastive learning. 
%that the negative samples closer to the positive samples, called hard negatives, contribute the most to the comparative learning while those negative samples far away from positive samples is irrelevant, and even too many such negative samples will damage the model. 
For those negative instances that are far away from the query instance $h_i^q$, the contributions of these nodes are rather limited as they could be easily distinguished w.r.t. $h_i^q$. However for those close negative instances, it is hard for the model to discriminate them and thus their contributions should be assigned with higher weights. %From the perspective of mutual information, it will learn the larger mutual information representation. Nevertheless, this does not mean that those remote nodes are useless at all. 

%In the initial training stage, the model does not have such a strong ability to discriminate negative samples. It needs some easy-to-identify nodes to optimize the model toward the right direction.
Accordingly, the weight of the $j$-th negative instance's embedding $h_j^k$ w.r.t. the $i$-th positive instance's embeddings $h_i^q$ is calculated as $w_{i,j}= {h_i^q \cdot h_j^k}/\tau_w$, where $\tau_w$ is a temperature hyper-parameter. Thus, the overall node-wise contrastive loss could be written as 

% It is mentioned in some papers\cite{2020hard} that the negative samples closer to the positive samples, called hard negatives, contribute the most to the comparative learning while those negative samples far away from positive samples is irrelevant, and even too many such negative samples will damage the model. The reason behind these results is these far away nodes are easily distinguishable nodes, so these negative samples have limited improvement in representation learning, and some negative samples with close distance can make nodes learn more unique representations. From the perspective of mutual information, it will learn the larger mutual information representation. Nevertheless, this does not mean that those remote nodes are useless at all. In the initial training stage, the model does not have such a strong ability to discriminate negative samples. It needs some easy-to-identify nodes to optimize the model toward the right direction. Therefore, we propose to give a weight for each negative sample. We set the weight of the negative sample $h_j^k$ sample for the query embedding $h_i^q$ is $w_{i,j}= {h_i^q \cdot h_j^k}/\tau_w$, where $\tau_w$ is a temperature hyper-parameter. Finally, the overall contrastive loss of our approach could be written as 
%-----------------------
\begin{equation}
{L}= -\sum_{i=1}^N \log \frac{e^{f({h}_i^q,{h}_i^k)}}{ e^{f({h}_i^q,{h}_i^k)}+\sum_{j \in S_i^-} w_{i,j}e^{f({h}_i^q,h_j^k)}}.
\end{equation}
%-----------------------
\section{Experimental Results}
\label{sec:exp}
In this section, we ﬁrst brieﬂy introduce experimental datasets, evaluation metrics as well as the experimental settings. Then, to evaluate the model performance, we not only compare ours method with unsupervised models, but also some supervised models to fully demonstrate the effectiveness of our approach. Extensive experiments are evaluated on several real-world datasets to %Finally, we further conduct model analysis, including ablation studies, qualitative analysis and sensitivity of hyper-parameters. Through experiment results and analysis, we will 
answer following research questions:

%-----------------------
\begin{itemize}
% [leftmargin=*]
\item \textbf{RQ1:} Whether the proposed approach %ASGCN 
outperforms the state-of-the-art supervised and unsupervised methods or not?
\item \textbf{RQ2:} Whether the proposed components could affect the model performance or not (ablation study)?
\item \textbf{RQ3:} Whether the proposed approach is sensitive to model parameters or not?
\item \textbf{RQ4:} The visualization results of the learnt item embeddings. 
\end{itemize}
%-----------------------

\subsection{Experimental Setup}
\subsubsection{Datasets}
In the experiments, three widely adopted real-world datasets are adopted to evaluate the model performance including 
%extracted from citation network are chosen which are 
$Cora$ \cite{cora}, $Citeseer$ \cite{citeseer} and $Pubmed$ \cite{pubmed}. %where documents (nodes) are connected through citations (edges). 
We follow the work \cite{2018gat} to partition each dataset into training set, validation set and test set. The statistics of these datasets are reported in Table \ref{tab:dataset}.

%----------------------------------------------------------------
\begin{table}[htbp]
\setlength{\abovecaptionskip}{5pt}
\setlength{\belowcaptionskip}{0pt}
\setlength{\tabcolsep}{5mm}
\centering
\begin{tabular}{ccccc}
\toprule[1pt]
\textbf{Dataset} & \textbf{\# of Nodes} & \textbf{\# of Edges} & \textbf{\# of Features} & \textbf{\# of Classes} \\
\midrule
Core & 2708 & 5429  & 1433  & 7 \\
Citeseer & 3327 & 4732 & 3703 & 6 \\
Pubmed & 19717 & 44338 & 500 & 3 \\
\bottomrule[1pt]
\end{tabular}
\caption{The statistics of experimental datasets.}\label{tab:dataset}
\end{table}
%----------------------------------------

\subsubsection{Baseline models}
To evaluate the model performance of the proposed approach on node classification task, both the unsupervised and supervised methods are compared in the experiments. 

%----------------------------------------------------------------
\begin{table}[t]
\setlength{\abovecaptionskip}{5pt}
\setlength{\belowcaptionskip}{0pt}
\setlength{\tabcolsep}{5mm}
\centering
\begin{tabular}{lllll}
\toprule[1pt]
\textbf{Available data} & \textbf{Method} & \textbf{Cora} & \textbf{Citeseer} & \textbf{Pubmed} \\
\midrule
\textbf{X},\textbf{Y}            & Raw features  & 55.1\%  & 46.5\%  & 71.4\% \\
\textbf{A},\textbf{Y}            & LP            & 68.0\%  & 45.3\%  & 63.0\% \\
\textbf{X},\textbf{A},\textbf{Y} & GCN           & 81.5\%  & 70.3\%  & 79.0\% \\
\textbf{X},\textbf{A},\textbf{Y} & Chebyshev     & 81.2\%  & 69.8\%  & 74.4\% \\
\textbf{X},\textbf{A},\textbf{Y} & GAT           & 83.0\%  & 72.5\%  & 79.0\% \\
\textbf{X},\textbf{A},\textbf{Y} & GeniePath     & 75.5\%  & 64.3\%  & 78.5\% \\
\textbf{X},\textbf{A},\textbf{Y} & JK-Net        & 82.7\%  & 73.0\%  & 77.9\% \\
\textbf{X},\textbf{A},\textbf{Y} & MixHop        & 81.9\%  & 71.4\%  & 80.8\% \\
\midrule
\textbf{A}                       & Deepwalk      & 67.2\%  & 43.2\%  & 65.3\% \\
\textbf{X},\textbf{A}            & Deepwalk+features& 70.7\%  & 51.4\%  & 74.3\% \\
\textbf{X},\textbf{A}            & GAE           & 71.5\%  & 65.8\%  & 72.1\% \\
\textbf{X},\textbf{A}            & GraphSAGE     & 68.0\%  & 68.0\%  & 68.0\% \\
\textbf{X},\textbf{A}            & DGI           & 82.3\%  & 71.8\%  & 76.8\% \\
\textbf{X},\textbf{S}            & DGI           & 83.8\%  & 72.0\%  & 77.9\% \\
\textbf{X},\textbf{A},\textbf{S} & Mvgrl         & \textbf{86.8\%}  & \underline{73.3\%}  & \underline{80.1\%} \\
\midrule
\textbf{X},\textbf{A}            & ours w/o all & 83.5\%  & 69.3\%  & 80.6\% \\

\textbf{X},\textbf{A}            & ours & \underline{84.3\%}  & \textbf{73.5\%}  & \textbf{81.5\%} \\
%ablation results
\bottomrule[1pt]
\end{tabular}
\caption{The average node classification results for both supervised and unsupervised models. The available data column highlights the data available to each model during the model training process (\textbf{X}:features, \textbf{A}:adjacency matrix, \textbf{S}:diffusion matrix, \textbf{Y}:labels). }
\label{tab:results}
\end{table}

%------------------------------------------------
\vspace{-0.1cm}
The \textbf{unsupervised} models we used in the experiment are as follows
\vspace{-0.1cm}
%-----------------------
\begin{itemize}
% [leftmargin=*]
\item \textbf{\emph{Deepwalk}} \cite{2014deepwalk} first samples related nodes via a truncated random walk, and then %employs negative sampling strategy to learn node level feature representations.
constructs negative examples to learn node embeddings. This approach is considered as the baseline method. We also compare our results to DeepWalk with the features concatenated, denoted as DeepWalk+features.

\item \textbf{\emph{GAE}} \cite{2016gae} is considered as the SOTA approach which applies variational auto-encoder to graphs firstly.

\item \textbf{\emph{GraphSAGE}} \cite{2018graphsage} is proposed to learn a function for generating low-dimensional embeddings by aggregating the embeddings of more informative neighbor nodes. % features from the local neighbors of a node, 
We use the unsupervised loss function mentioned in \cite{2018graphsage} to train the model.
% This method is an inductive one and thus can embed unseen nodes. We use the unsupervised loss function mentioned in \cite{2018graphsage} to train the model.

\item \textbf{\emph{DGI}} \cite{2018DGI} is considered as the SOTA unsupervised learning approach which maximizes the mutual information between the node-level and the graph-level feature embeddings.

\item \textbf{\emph{Mvgrl}} \cite{2020multi-view} is the SOTA self-supervised  method proposed to %learning both node- and graph-level representations by contrasting structural views for graphs.
learn node level embeddings by optimizing the contrast between node representations and view-level graph representations. 
\end{itemize}
%-----------------------

The \textbf{semi-supervised} models we used in the experiment are as follows
\vspace{-0.1cm}
%-----------------------
\begin{itemize}
\item \textbf{\emph{LP}} \cite{2003LP} assigns labels to unlabeled samples and is considered as a baseline method. %, and the core idea is that similar nodes should have the same label.

\item \textbf{\emph{GCN}} \cite{2017GCN} is one of the milestone GNN models originally proposed for node classiﬁcation problem. And the model is trained by minimizing the supervised loss.

\item \textbf{\emph{Chebyshev}} \cite{2017chebyshev} designs the convolution kernels using the Chebyshev inequality to speed up the Fourier transformation for the graph convolution process, and is considered as the baseline.

% \item \textbf{\emph{GraphSAGE-GCN}} \cite{2018graphsage} applies GCN to inductively aggregate feature embeddings of neighboring nodes.

\item \textbf{\emph{GAT}} \cite{2018gat} is essentially an attention based approach. GAT designs a multi-head self-attention layer to assign weights to different feature embeddings of graph nodes, and is also treated as the baseline.

\item \textbf{\emph{GeniePath}} \cite{2018geniepath} samples neighboring nodes which contribute a lot to the target node via a hybrid of BFS and DFS search strategy. % breadth ﬁrst search and depth ﬁrst search strategy, respectively.

\item \textbf{\emph{JK-Net}} \cite{2018JK-Net} adaptively uses different neighborhood ranges for each node to perform aggregation operations.

\item \textbf{\emph{MixHop}} \cite{2019mixhop} proposes to perform multi-order convolution to learn general mixing of neighborhood information.

% \item \textbf{\emph{MoNet}} \cite{2020monet} is considered as the SOTA approach which designs an elegant model to incorporate motion features into the predictions.
\end{itemize}
%-----------------------
\vspace{-0.3cm}
\subsubsection{Setting of Model Parameters}
Our task is to first train the node embeddings and then directly evaluate its node classification ability. We set the same experimental settings as the SOTA \cite{2018DGI,2020multi-view} and report the mean classiﬁcation results on the testing set after 50 runs of training followed by a linear model.
% For both node classiﬁcation datasets, we evaluate the proposed approach under the linear evaluation protocol and we closely follow the experimental protocol of the previous state-of-the-art approaches \cite{2018DGI}\cite{2020multi-view} and report the mean classiﬁcation accuracy on the test nodes after 50 runs of training followed by a linear model. 
We initialize the parameters using Xavier initialization \cite{Xavier} and train the model using Adam optimizer \cite{2017adam} with an initial learning rate of 0.001. We follow the same settings as DGI does and set the number of epochs to 2000. We vary the the batch size from 50 to 2000, %And then we choose the batch size from [50,2000]. 
and the early stopping with a patience of 20 is adopted. The embedding dimension is set to 512. Unlike DGI, we use two layers of GCN. We set the step of random walk as 25, soft-margin $\alpha$ as 0.9, dropout rate as 0.7.
%Finally, we set the size of hidden dimension of node embeddings to 512. For chosen hyper-parameters see Appendix.

%ablation experiment

\subsection{RQ1: Performance Comparison}
In this experiment, we first learn the node embeddings and then use these embeddings to directly evaluate the node classification task, and the results are reported in Table \ref{tab:results}. Obviously, the proposed approach achieves the best results both in comparison with unsupervised models or semi-supervised models, except for Cora dataset where the Mvgrl achieves the best results, and ours is the second best one. Particularly, the accuracy on Pubmed model, which has the most nodes, is improved by 81.5\%. As the Mvgrl method could make full use of global diffusion information, the model performance is expected to be superior to ours. But due to the use of diffusion information, Mvgrl cannot be used in inductive way, which limits its applicability. However, it is well noticed that our approach is better than the SOTA DGI trained with $S$. This verifies the effectiveness of our approach. It is also noteworthy that our model has already performed well on the Cora and Pubmed datasets without adding any modules, that is, only applying node-level comparison between nodes.
% show that we achieve state-of-art results in two of the three sets with respect to previous unsupervised and semi-supervised models, and our performance on the $Cora$ is second only to Mvgrl. The reason why Mvgrl performs well may be attributed to the use of diffusion matrix $\textbf{S}$, which has been proved to be useful for graph representation learning\cite{klicpera2019diffusion}. However, the calculation of \textbf{S} needs to use global adjacency information, which is not available in the inductive scenario. That is, the application of Mvgrl is limited.

% \blue{To evaluate node classification under the clustering protocol}

\subsection{RQ2: Ablation Study}
In this experiment, we investigate the effectiveness of the proposed component. We respectively remove the component of soft-margin sampling, DPP sampling and node weights, and report the results in Table \ref{tab:ablation results}. %We carry out thorough ablation studies on the three modules, negative sampling, dpp sampling and node weights. Here, we report the results in Table \ref{tab:ablation results}.

%------------------------------------------------
\begin{table}[htbp]
\setlength{\abovecaptionskip}{5pt}
\setlength{\belowcaptionskip}{0pt}
\setlength{\tabcolsep}{5mm}
\centering
\begin{tabular}{lllll}
\toprule[1pt]
\textbf{Variants} &  \textbf{Cora} & \textbf{Citeseer} & \textbf{Pubmed} \\
\midrule
ours w/o all   & 83.5\%  & 69.3\%  & 80.6\% \\
ours with $\alpha$   & 83.8\%  & 70.9\%  & 80.8\% \\
ours with DPP  & 83.9\%  & 71.8\%  & 81.2\% \\
ours with $w$    & 83.8\%  & 70.1\%  & 80.9\% \\
ours          & \textbf{84.3\%}  & \textbf{73.5\%}  & \textbf{81.5\%} \\
%ablation results
\bottomrule[1pt]
\end{tabular}
\caption{The ablation study results. In this table, ours w/o all denotes that we remove all proposed components. %model that only apply node-level contrastive strategy. 
And ours with $\alpha$, ours with DPP, ours with $w$ denote the model with soft-margin sampling, DPP sampling and node weights, respectively.}
\label{tab:ablation results}
\end{table}
%-----------------------------------------------------------------
\vspace{-1.2cm}
\subsubsection{Effect of node weight $w$} 
From Table \ref{tab:ablation results}, it is noticed that if removing all components, the model performance is the worst. If the model is integrated with node weight, the model performance slightly increases. This indicates that the contribution of node weight is not significant. The reason may be the weight function we designed is too simple.  
%For probe the ability of proposed node weight mechanism, we confer different weight $w$ for each node. In particular, we give higher weight to negative samples which are more similar with each other, give lower weight otherwise, the detailed hyper-parameter settings are same as experiments in XX. The results in XX show our module of node weight improve the model's performance, compared with the vanilla model whose valid information might be drown by redundant information under undifferentiated weight mechanism, which indicates our node weight mechanism is satisfactorily  effective.

\vspace{-0.3cm}
\subsubsection{Effect of distribution of $p_i^+$ and $p_i^-$}
It is noticed that the model performance of ``ours with $\alpha$'', i.e., the positive and negative instance sets are generated by the learnt data distribution, improves. This partially verifies the effectiveness of the proposed component. 
%TODO: introduce the experiment. The results in XX demonstrate that the model's performance increases with our proposed module, which suggests the proposed model indeed contribute to choose positive samples from negative samples set ,and thus mitigate class collision of unsupervised contrastive learning.
\vspace{-0.8cm}
\subsubsection{Effect of DPP sampling}
It could be observed that the ``ours with DPP'' achieves the second best results w.r.t. all evaluation criteria. This verifies our proposed assumption that the data distribution of negative examples is a key factor in affecting the model performance. 

% similar observations in \cite{2020hard} that the number of negative instances is a key factor affecting the %quality of the learnt embeddings, and this consequently affects the model performance. 
\vspace{-0.3cm}
\subsection{RQ3: Parameter Analysis}% Sensitivity of hyper-parameters}
\begin{figure}[htb]
\setlength{\abovecaptionskip}{5pt}
\setlength{\belowcaptionskip}{0pt}
\centering
\subfigure[Results on batch size] {
\includegraphics[width=1.4in]{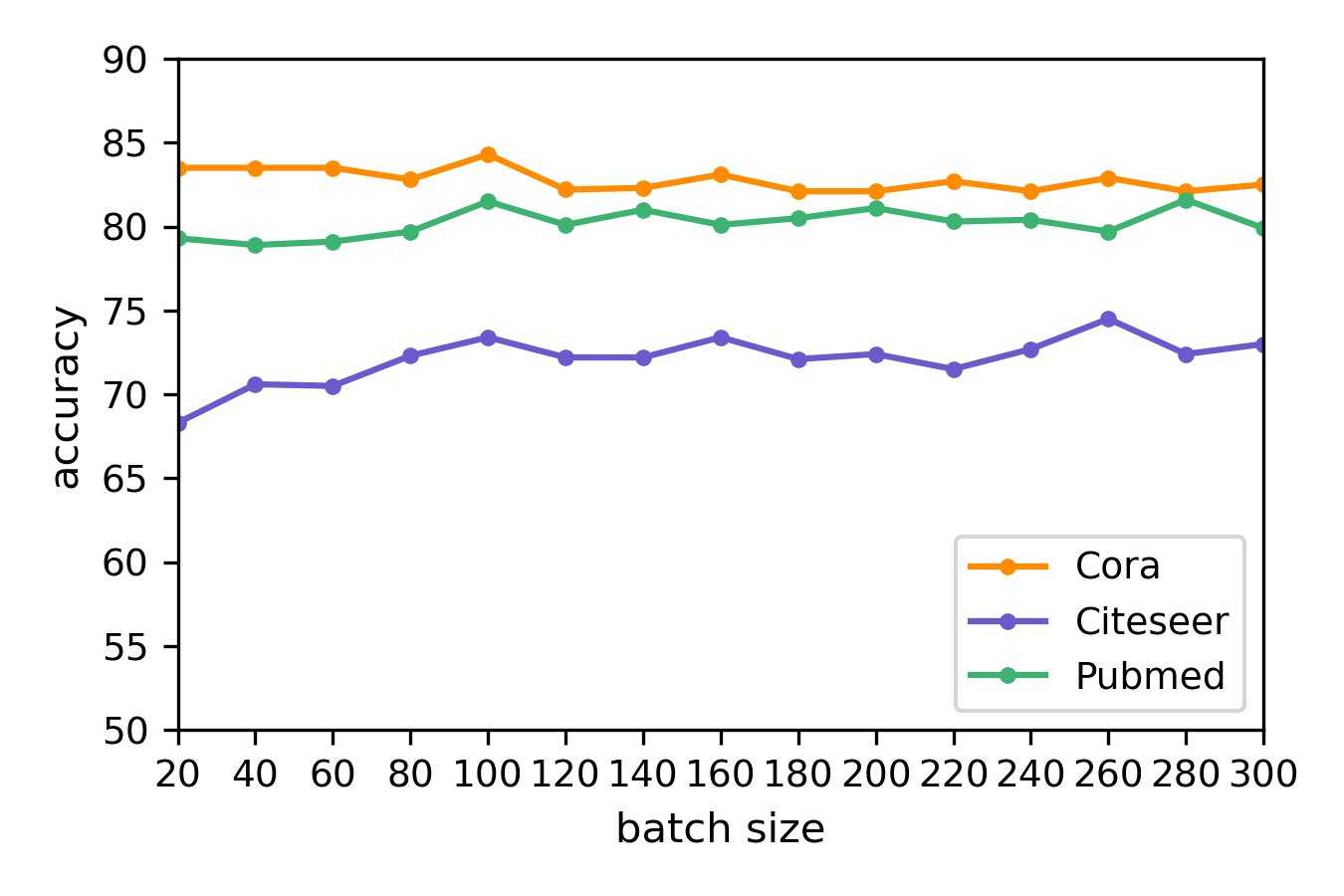}
% \subfigcapskip = 50pt
\label{fig:parameter_1}}
\quad
\subfigure[Results on random walk length]{
\includegraphics[width=1.4in]{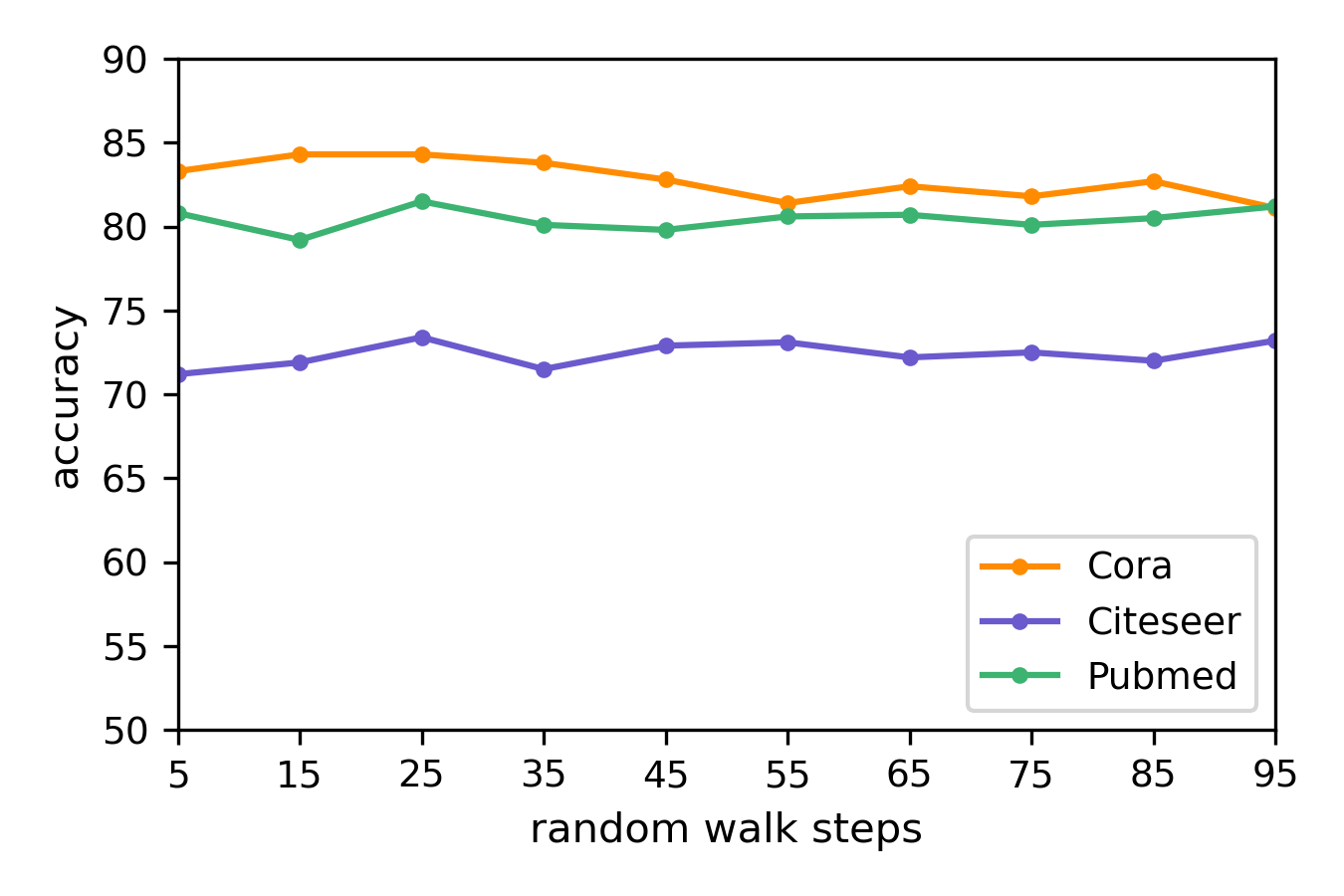}
\label{fig:parameter_2}}
\quad
\subfigure[Results on soft-margin $\alpha$]{
\includegraphics[width=1.4in]{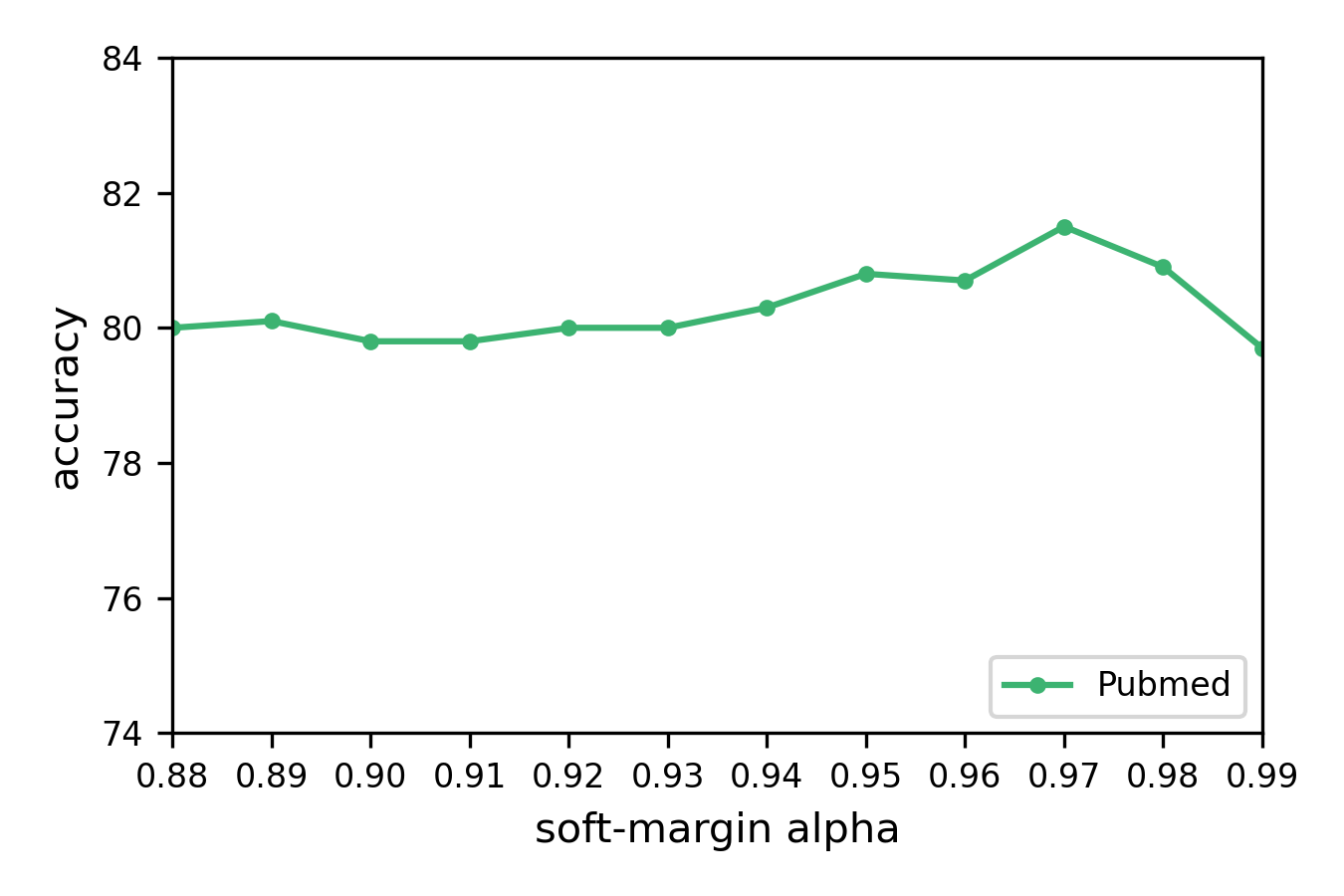}
\label{fig:parameter_3}}
\caption{Parameter analysis results.}
\label{fig:hyper-parameter}
\end{figure}

In the section, we evaluate how the model parameters, e.g., the step of random walk, soft-margin $\alpha$ and batch size, affect the model performance, and the corresponding results are plotted in Figure \ref{fig:hyper-parameter}. 

% --------------------
% ----------------

%Here, we give sufficient experiments and analyse the effect of size of subgraph, strictness parameter $\alpha$, rate of DPP and batch size, respectively. 
From this figure, we have following observations. First, we highlight that our model is insensitive to parameter ``batch size'' and ``random walk length '', as shown in Fig \ref{fig:parameter_1} and \ref{fig:parameter_2}. Second, our model is obviously sensitive to the soft-margin parameter $\alpha$ which controls the ratio of  of selecting potential positive samples from the negative samples set . It is also well noticed that when $\alpha$ is close to 1, the model performance dramatically drops. This verifies that negative samples sampling is crucial to the model performance of the contrastive learning. 

% with the number of positive examples is also crucial to the model performance of the contrastive learning. 

% We highlight that our model is insensitive to these hyper-parameters, which enable our model to be robust to different hyper-parameters and further avoid practitioner manually adjust these hyper-parameter values for better performance. Specifically, result in xx show that though the performance varies when hype-parameter settings are extreme, otherwise there is marginal effect on performance for different hype-parameter settings in certain extent.

% \vspace{-0.4cm}
% \subsubsection{step of random walk}
% The result in xx show that although when the size of subgraph transform into extremely low or high, the performance moves sharply. However, the performance change marginally as the size of subgraph gradually varies.
% \vspace{-0.4cm}
% \subsubsection{strictness
% parameter $\alpha$}
% Similarly, for strictness value $\alpha$ in certain range, there is little effect on the performance. The xxx vary sharply once strictness parameter is  beyond the extent, which is reasonable and mainly relates to the incorporation of samples. When the strictness is too relax(strictness value $\alpha$ is too small), a amount of negative samples are introduced into model which harm the the model's performance; However giving a large  strictness parameter $\alpha$(too strict), rarely positive samples are injected, therefore there is no evident improvement on representation of node.
% \vspace{-0.4cm}

% --------------------
\begin{figure}[htb]
\setlength{\abovecaptionskip}{5pt}
\setlength{\belowcaptionskip}{0pt}
\centering
\subfigure[Raw features] {
\includegraphics[width=1.4in]{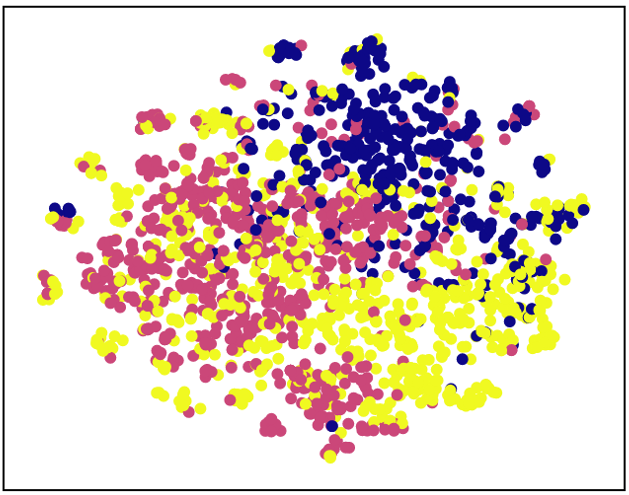}
% \subfigcapskip = 50pt
\label{fig:qua_1}}
\quad
\subfigure[Mvgrl]{
\includegraphics[width=1.4in]{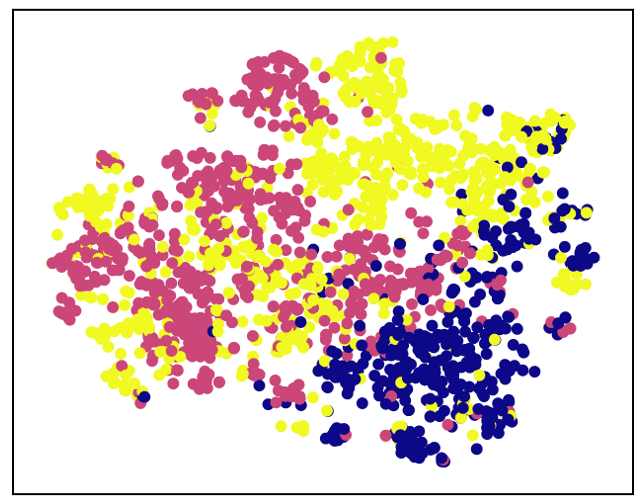}
\label{fig:que_2}}
\quad
\subfigure[Ours]{
\includegraphics[width=1.4in]{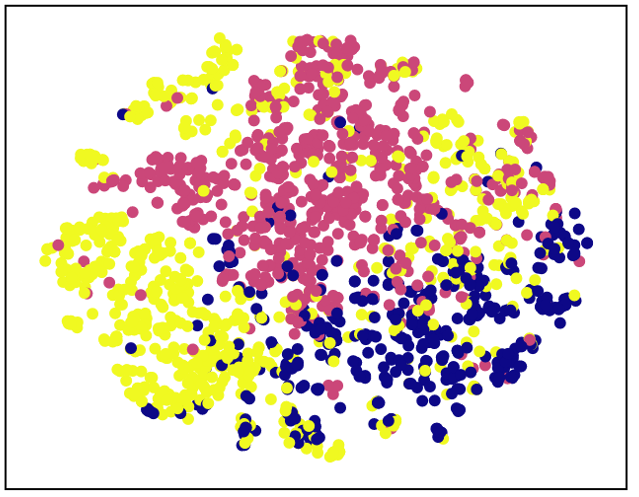}
\label{fig:que_3}}
\caption{Visualizing the learnt embeddings of nodes on Pubmed dataset. In this figure, each color represents one underlying true class, and each colored point represent the embeddings of a node. % \textbf{(a)} Raw features, \textbf{(b)} a learned DGI model, \textbf{(c)} a learned Mvdgl and \textbf{(d)} a learned our model.
}
\label{fig:quelitative analysis}
\end{figure}
% -----------------
\vspace{-0.3cm}
\subsection{RQ4: Visualization Results}
Due to page limitations, we choose to visualize the node embeddings learnt on Pubmed dataset to provide a vivid illustration of the proposed model performance. As there are many nodes in Pubmed, we randomly selected 2000 nodes to plot. We also visualized the results of the Mvgrl method for comparison as this approach achieves the superior node classification results. The visualization results are plotted in Figure \ref{fig:quelitative analysis}. Obviously, our model achieves the best visualization results on Pubmed dataset. It is noticed that in the raw features, most of the ``red'' and ``yellow'' class are mixed up together, which makes the classification task difficult. It is also noticed that the three classes could be well separated and spread over the whole data space by our model, whilst the ``red'' class and ``yellow'' class are still mixed up in Figure \ref{fig:que_2}. We can infer that our approach could verifies that node-level contrastive learning can learn informative low-dimensional representations. 
\vspace{-0.2cm}
\section{Conclusion}
\label{sec:conc}
\vspace{-0.1cm}

In this paper, we propose a novel node-wise contrastive learning approach to learn node embeddings for a supervised task. Particularly, we propose to resolve class collision issue by transiting the detected ``in doubt'' negative instances from the negative instance set to the positive instance set. Furthermore, a DPP-based sampling strategy is proposed to evenly sample negative instances for the contrastive learning. Extensive experiments are evaluated on three real-world datasets and the promising results demonstrate that the proposed approach is superior to both the baseline and the SOTA approaches. 

\newpage
\bibliographystyle{splncs04}
\bibliography{myref}
%
%\begin{thebibliography}{8}
%\bibitem{ref_article1}
%Author, F.: Article title. Journal \textbf{2}(5), 99--110 (2016)

%\bibitem{ref_lncs1}
%Author, F., Author, S.: Title of a proceedings paper. In: Editor,
%F., Editor, S. (eds.) CONFERENCE 2016, LNCS, vol. 9999, pp. 1--13.
%Springer, Heidelberg (2016). \doi{10.10007/1234567890}

%\bibitem{ref_book1}
%Author, F., Author, S., Author, T.: Book title. 2nd edn. Publisher,
%Location (1999)

% \bibitem{ref_proc1}
% Author, A.-B.: Contribution title. In: 9th International Proceedings
% on Proceedings, pp. 1--2. Publisher, Location (2010)

% \bibitem{ref_url1}
% LNCS Homepage, \url{http://www.springer.com/lncs}. Last accessed 4
% Oct 2017
% \end{thebibliography}
\end{document}